\documentclass[11pt,a4paper]{article}

\usepackage{url}            
\usepackage{booktabs}       
\usepackage{amsfonts}       
\usepackage{nicefrac}       
\usepackage{microtype}      
\usepackage{graphicx}
\usepackage{amsmath,amssymb} 
\usepackage{array}
\usepackage{bbding} 
\usepackage{paralist}
\usepackage{xspace}
\usepackage{subcaption}
\usepackage{xcolor} 
\usepackage{mathtools}
\usepackage{comment}

\definecolor{cadmiumgreen}{rgb}{0.0, 0.42, 0.24}
\definecolor{amethyst}{rgb}{0.6, 0.4, 0.8}
\definecolor{magentagoal}{RGB}{255, 0, 255}
\definecolor{cyangoal}{RGB}{60, 255, 255}

\newcolumntype{L}[1]{>{\raggedright\let\newline\\\arraybackslash\hspace{0pt}}m{#1}}
\newcolumntype{C}[1]{>{\centering\let\newline\\\arraybackslash\hspace{0pt}}m{#1}}
\newcolumntype{R}[1]{>{\raggedleft\let\newline\\\arraybackslash\hspace{0pt}}m{#1}}
\newcolumntype{H}[1]{>{\raggedright\let\newline\\\arraybackslash\hspace{0pt}}m{#1}}

\DeclareMathSymbol{@}{\mathord}{letters}{"3B}

\newcommand{\archi}{\texttt{E2E-PExt$_\texttt{E}$}\xspace}

\newcommand{\xia}{\texttt{ECPE 2-stage}\xspace}
\newcommand{\dingone}{\texttt{ECPE-2D(BERT)}\xspace}

\newcommand{\dingtwoisml}{\texttt{ECPE-MLL(ISML-6)}\xspace}

\newcommand{\rhnn}{\texttt{RHNN}\xspace}

\newcommand{\bound}{\texttt{E2E-CExt}\xspace}

\newcommand{\ce}{\texttt{E2E-PExt$_\texttt{C}$}\xspace}

\newcommand{\cebound}{\texttt{E2E-EExt}\xspace}
\usepackage[hyperref]{eacl2021}
\usepackage{times}
\usepackage{latexsym}

\usepackage{graphicx}
\usepackage{arydshln}
\usepackage[skip=0.3\baselineskip]{caption}
\usepackage{wrapfig}

\usepackage[compact]{titlesec}

\usepackage{microtype}

\aclfinalcopy 

\usepackage{lipsum}
\usepackage{mathtools}
\usepackage{cuted}

\title{An End-to-End Network for Emotion-Cause Pair Extraction}

\author{Aaditya Singh\footnotemark[1]\thanks{\quad Currently affiliated with IBM Research} \footnotemark[2] \qquad    
  Shreeshail Hingane\footnotemark[2]\qquad
  Saim Wani\thanks{\quad Authors equally contributed} \qquad
  \large{\textbf{Ashutosh Modi}} \\
{Indian Institute of Technology Kanpur (IIT Kanpur)} \\
  {\tt \{aadisin, shreessh, saim\}@iitk.ac.in}  \\
  {\tt ashutoshm@cse.iitk.ac.in}  \\
}
  
\date{}

\begin{document}
\maketitle
\begin{abstract}

The task of Emotion-Cause Pair Extraction (ECPE) aims to extract all potential clause-pairs of emotions and their corresponding causes in a document. Unlike the more well-studied task of Emotion Cause Extraction (ECE), ECPE does not require the emotion clauses to be provided as annotations. Previous works on ECPE have either followed a multi-stage approach where emotion extraction, cause extraction, and pairing are done independently or use complex architectures to resolve its limitations. In this paper, we propose an end-to-end model for the ECPE task. 
Due to the unavailability of an English language ECPE corpus, we adapt the NTCIR-13 ECE corpus and establish a baseline for the ECPE task on this dataset. On this dataset, the proposed method produces significant performance improvements ($\sim 6.5\%$ increase in F1 score) over the multi-stage approach and achieves comparable performance to the state of the art methods. 

\end{abstract}

\begin{figure*}[t]
\begin{center}
\begin{tabular}{ :rl: } 
 \hdashline
\texttt{Clause 1:} & Adele arrived at her apartment late in the afternoon after a long day of work.  \\
\texttt{Clause 2:} & She was still furious with her husband for not remembering her 40th birthday.  \\ 
\texttt{Clause 3:} & As soon as she unlocked the door, she gasped with surprise; \\ 
\texttt{Clause 4:} & Mikhael and Harriet had organized a huge party for her.  \\
 \hdashline
\end{tabular}
\end{center}
\caption{An example document. The above example contains two emotion-cause pairs. Clause 2 is an emotion clause (\textit{furious}) and is also the corresponding cause clause (\textit{for not remembering her 40th birthday}). Clause 3 is an emotion clause (\textit{surprise}) and Clause 4 is its corresponding cause clause (\textit{organized a huge party for her}).}
\label{fig:ex}
\end{figure*}

\section{Introduction}
\label{intro}

There have been several works on emotions prediction from the text \citep{alswaidan2020survey, witon-etal-2018-disney} as well as generating emotion oriented texts \citep{ghosh2017affect, colombo-etal-2019-affect, amodi-affect-dialog-patent-2020, goswamy-etal-2020-adapting}. However, recently the focus has also shifted to finding out the underlying cause(s) that lead to the emotion expressed in the text. 
In this respect, \citet{gui-etal-2016-event} proposed the Emotion Cause Extraction (ECE), a task aimed at detecting the cause behind a given emotion annotation. The task is defined as a clause level classification problem. The text is divided at the clause level and the task is to detect the clause containing the cause, given the clause containing the emotion. However, the applicability of models solving the ECE problem is limited by the fact that emotion annotations are required at test time.  More recently, \newcite{xia2019emotion} introduced the Emotion-Cause Pair Extraction (ECPE) task i.e. extracting all possible emotion-cause clause pairs in a document with no emotion annotations. Thus, ECPE opens up avenues for applications of real-time sentiment-cause analysis in tweets and product reviews. ECPE builds on the existing and well studied ECE task. Figure \ref{fig:ex} shows an example with ground truth annotations.
\par \newcite{xia2019emotion} use a two-stage architecture to extract potential emotion-cause clauses. In Stage 1, the model extracts a set of emotion clauses and a set of cause clauses (not mutually exclusive) from the document. In Stage 2, it performs emotion-cause pairing and filtering, i.e. eliminating pairs that the model predicts as an invalid emotion-cause pair. However, this fails to fully capture the mutual dependence between emotion and cause clauses since clause extraction happens in isolation from the pairing step. Thus, the model is never optimized using the overall task as the objective. Also, certain emotion clauses are likely not to be detected without the corresponding cause clauses as the context for that emotion. Recent methods such as \citet{ding2020ecpe} and \citet{ding2020end} use complex encoder and classifier architectures to resolve these limitations of the multi-stage method.
\par In this paper, we propose an end-to-end model to explicitly demonstrate the effectiveness of joint training on the ECPE task. The proposed model attempts to take into account the mutual interdependence between emotion and cause clauses. 
Based on the benchmark English-language corpus used in the ECE task of the NTCIR-13 workshop \cite{gaooverview}, we evaluate our approach on this dataset after adapting it for the ECPE task. 
We demonstrate that the proposed approach works significantly better than the multi-stage method and achieves comparable performance to the state of the art methods. We also show that when used for the ECE task by providing ground truth emotion annotations, our model beats the state of the art performance of ECE models on the introduced corpus. We provide the dataset and implementations of our models via GitHub\footnote{\url{https://github.com/Aaditya-Singh/E2E-ECPE}}.

\section{Related Work}
The problem of Emotion Cause Extraction (ECE) has been studied extensively over the past decade. 
ECE was initially proposed as a word-level sequence detection problem in \citet{lee-etal-2010-text}. Attempts to solve this task focused on either classical machine learning techniques \cite{ghazi2015detecting}, or on rule-based methods  \cite{Neviarouskaya2013ExtractingCO,10.1016/j.eswa.2015.01.064}. Subsequently, the problem was reframed as a clause-level classification problem \cite{chen-etal-2010-emotion} and the Chinese-language dataset introduced by \newcite{gui-etal-2016-event} has since become the benchmark dataset for ECE and the task has been an active area of research \cite{xia2019rthn,gui-etal-2017-question,8598785,li-etal-2018-co,Li2019ContextawareEC,fan-etal-2019-knowledge}. 
\par However, the main limitation of ECE remains that it requires emotion annotations even during test time, which severely limits the applicability of ECE models. To address this, \newcite{xia2019emotion} introduced a new task called emotion-cause pair extraction (ECPE), which extracts both emotion and its cause without requiring the emotion annotation. They demonstrated the results of their two-stage architecture on the benchmark Chinese language ECE corpus \cite{gui-etal-2016-event}. Following their work, several works have been proposed to address the limitations of the two-stage architecture (\newcite{ding2020ecpe}, \newcite{ding2020end}, \newcite{fan2020transition}, \newcite{yuan2020emotion}, \newcite{cheng2020symmetric}, \newcite{chen2020unified}) . In order to explore the corpus further and to encourage future work from a broader community, we use an English language ECE corpus. (see section \ref{sec:corpus}). 

\section{Task Definition}
Formally, a document consists of text that is segmented into an ordered set of clauses $\mathbf{\text{D}} = [c_1, c_2, ..., c_{d}]$ and the ECPE task aims to extract a set of emotion-cause pairs $\mathbf{\text{P}} = \{...,(c_i, c_j),...\}$ ($c_i, c_j \in \mathbf{\text{D}}$), where $c_i$ is an emotion clause and $c_j$ is the corresponding cause clause. In the ECE task, we are additionally given the annotations of emotion clauses and the goal is to detect (one or more) clauses containing the cause for each emotion clause.

\begin{figure*}[t]
\centering
\includegraphics[scale=0.75]{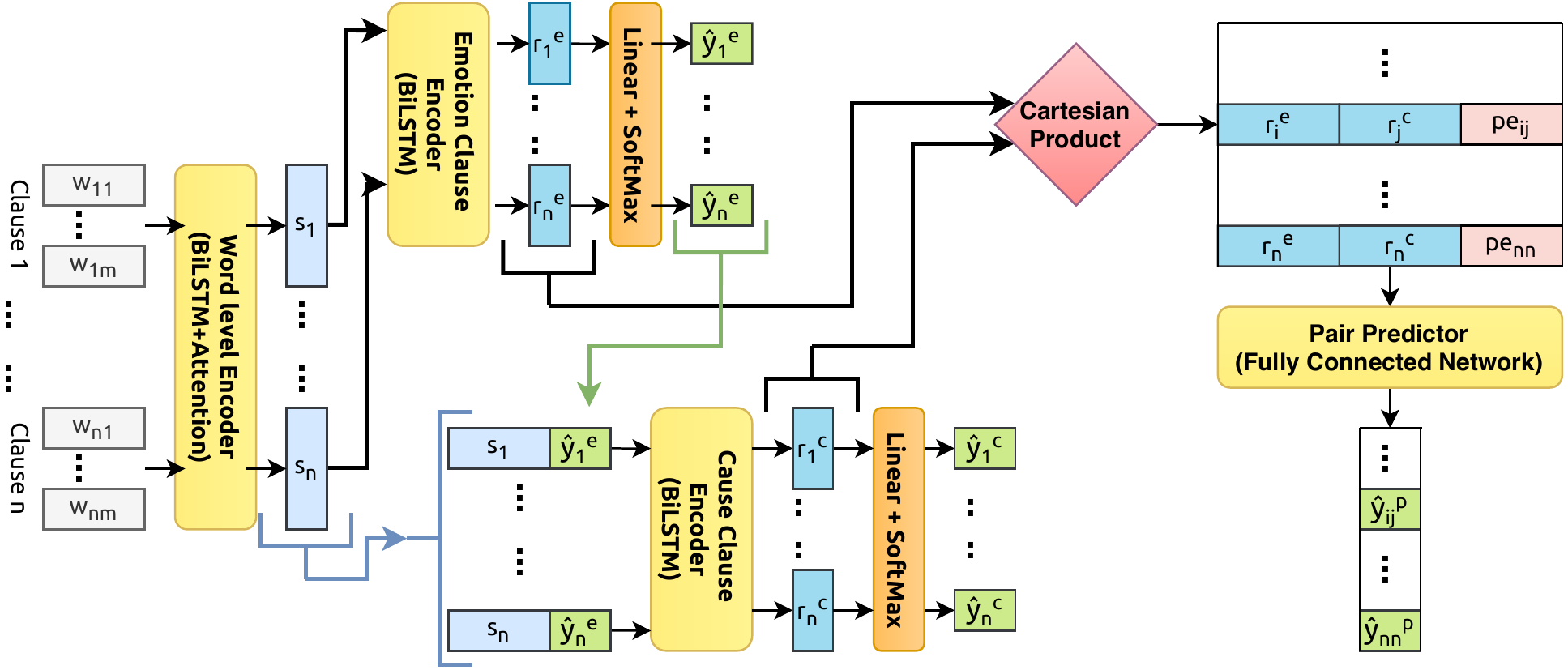}
\caption{End-to-End network (\archi)}
\label{fig:E2E}
\end{figure*}

\section{Approach}
We propose an end-to-end emotion cause pairs extraction model (Figure \ref{fig:E2E}), henceforth referred to as  \archi (refer to section \ref{sec:experiments} for the naming convention). 
The model takes an entire document as its input and computes, for each ordered pair of clauses $(c_i, c_j)$, the probability of being a potential emotion-cause pair. To facilitate the learning of suitable clause representations required for this \textit{primary} task, we train the model on two other \textit{auxiliary} tasks: Emotion Detection and Cause Detection.

We propose a hierarchical architecture. Word level representations are used to obtain clausal representations (vi BiLSTM) and clause level representations are further \textit{contextualized} using another BiLSTM network. The resulting contextualized clause representations are then used for the classification task. 
Let $\mathbf{w}_{i}^{j}$ represent the vector representation of the $j^{th}$ word in the $i^{th}$ clause. Each clause $\mathbf{c}_{i}$ in the document $d$ is passed through a \textit{word-level} encoder (BiLSTM + Attention \cite{bahdanau2014neural}) to obtain the clause representation $\mathbf{s}_{i}$. The clause embeddings are then fed into two separate \textit{clause-level} encoders (\texttt{Emotion-Encoder} and \texttt{Cause-Encoder}) each of which corresponds, respectively, to the two auxiliary tasks. The purpose of the clause-level encoders is to help learn \textit{contextualized} clause representations by incorporating context from the neighboring clauses in the document. For each clause $c_{i}$, we obtain its contextualized representations $\mathbf{r}_{i}^{e}$ and $\mathbf{r}_{i}^{c}$ by passing it through a BiLSTM network.   

\begin{table*}[t]
\centering
\label{results table}
\resizebox{\linewidth}{!}{
\begin{tabular}{@{} l rrr rrr rrr @{} }
\toprule
 & \multicolumn{3}{c}{Emotion Extraction} & \multicolumn{3}{c}{Cause Extraction} & \multicolumn{3}{c}{Pair Extraction}\\ \cmidrule(lr){2-4} \cmidrule(lr){5-7} \cmidrule(l{0mm}r){8-10} 
 & Precision  & Recall  & F1 Score  & Precision  & Recall & F1 Score  & Precision & Recall & F1 Score \\
\midrule
  \xia \cite{xia2019emotion} & 67.41 & 71.60 & 69.40 & 60.39 & 47.34 & 53.01 & 46.94 & 41.02 & 43.67  \\
  \dingone \cite{ding2020ecpe} & 74.35 & 69.68 & 71.89 & 64.91 & 53.53 & 58.55 & 60.49 & 43.84 & 50.73  \\
  \dingtwoisml \cite{ding2020end} & 75.46 & 69.96 & 72.55 & 63.50 & 59.19 & 61.10 & 59.26 & 45.30 & 51.21  \\
  \archi (Ours) & 71.63 & 67.49 & 69.43 & 66.36 & 43.75 & 52.26 & 51.34 & 49.29 & 50.17\\
\midrule
\rhnn \cite{fan-etal-2019-knowledge} & - & - & - & - & - & - & 69.01 & 52.67 & 59.75\\ 
\bound (Ours) & - & - & - & - & - & - & 65.21 & 66.18 & \textbf{65.63}\\
\bottomrule
\end{tabular}
}

\caption{\footnotesize{Experimental results comparing our models with the already existing benchmarks. The top half compares our model for ECPE (\archi) against the existing benchmarks. The bottom half compares existing ECE benchmark on this dataset against our model (\bound). Note that the Pair Extraction task in ECPE with true-emotion provided reduces to the Cause Extraction task of ECE. The results on \rhnn and \bound are only for the primary task, since in the ECE setting, there are no auxiliary tasks. The evaluation metrics are same as the ones used in previous works.}}
\label{tab:results}
\end{table*}

For the clause $c_{i}$, its contextual representations $\mathbf{r}_{i}^{e}$ and $\mathbf{r}_{i}^{c}$ are then used to predict whether the clause is an emotion-clause and a cause-clause respectively, i.e.,
\begin{align}
     & \mathbf{\hat{y}}_{i}^{e} = \text{softmax}(\mathbf{W}^{e}*\mathbf{r}_{i}^{e} + \mathbf{b}^{e});  \\
     & \mathbf{\hat{y}}_{i}^{c} = \text{softmax}(\mathbf{W}^{c}*\mathbf{r}_{i}^{c} + \mathbf{b}^{c})
\end{align}
As observed by \newcite{xia2019emotion}, we also noticed that performance on the two auxiliary tasks could be improved if done in an interactive manner rather than independently. Hence, the \texttt{Cause-Encoder} also makes use of the corresponding emotion-detection prediction $\mathbf{y}_{i}^{e}$, when generating $\mathbf{r}_{i}^{c}$ (Figure \ref{fig:E2E}).

For the primary task, every ordered pair $(c_i,c_j)$ is represented by concatenating $\mathbf{r}_{i}^{e}$, $\mathbf{r}_{j}^{c}$ and $\mathbf{pe}_{ij}$, wherein $\mathbf{pe}_{ij}$ is the positional embedding vector representing the relative positioning between the two clauses $i,j$ in the document  \cite{shaw2018self}. 

The primary task is solved by passing this pair-representation through a fully-connected neural network to get the pair-predictions. 
\begin{align}
    & \mathbf{r}_{ij}^{p} = [\mathbf{r}_{i}^{e} \oplus \mathbf{r}_{j}^{c} \oplus \mathbf{pe}_{ij}];\\
    & \mathbf{h}_{ij}^{p} = \text{ReLU}(\mathbf{W}^{p_{1}}*\mathbf{r}_{ij}^{p} + \mathbf{b}^{p_{1}});\\
    & \mathbf{\hat{y}}_{ij}^{p} = \text{softmax}(\mathbf{W}^{p_{2}}*\mathbf{h}_{ij}^{p} + \mathbf{b}^{p_{2}})
\end{align}

To train the end-to-end model, loss function is set as the weighted sum of loss on the primary task as well as the two auxiliary tasks i.e. $L^{\text{total}} = \lambda_{c}*L^{c} + \lambda_{e}*L^{e} + \lambda_{p}*L^{p}$, 
where $L^{e}$, $L^{c}$, and $L^{p}$ are cross-entropy errors for emotion, cause and pair predictions respectively. Further, $L^{p} = L^{pos} + loss\text{\_}weight*L^{neg}$, where $L^{pos}$ and $L^{neg}$ are the errors attributed to positive and negative examples respectively. We use hyperparameter $loss$\_$weight$ to scale down $L^{neg}$, since there are far more negative examples than positive ones in the primary pairing task.


\section{Corpus/Data Description} \label{sec:corpus}
We adapt an existing Emotion-Cause Extraction (ECE) \cite{fan-etal-2019-knowledge,Li2019ContextawareEC} corpus for evaluating our proposed models (as well as the architectures proposed in previous work). 
The corpus was introduced in the NTCIR-13 Workshop  \cite{gaooverview} for the ECE challenge.
\par The corpus consists of 2843 documents taken from several English novels. Each document is annotated with the following information: i) emotion-cause pairs present in the document, that is, the set of emotion clauses and their corresponding cause clauses; ii) emotion category of each clause; and iii) the keyword within the clause denoting the labeled emotion. We do not use the emotion category or the keyword during training of either ECE or ECPE tasks — only the emotion-cause pairs are used. At test time, none of the annotations are used for the ECPE task. For ECE task, emotion annotation is provided at test time and the model predicts the corresponding cause clauses. $80\%$-$10\%$-$10\%$ splits are used for training, validation and testing. 10 such randomly generated splits are used to get statistically significant results, and the average results are reported. 

\section{Experiments and Results} \label{sec:experiments}

\textbf{Naming Scheme for Model Variants:}\\
\texttt{E2E: } End to End.\\
\texttt{\texttt{PExt/CExt/EExt}}: Pair Extraction/ Cause Extraction/ Emotion Extraction.\\
Subscript \texttt{E/C:} represents how the auxiliary tasks are solved interactively (emotion predictions used for cause detection or vice versa).

For the purpose of evaluation on our dataset, we reproduced the two-stage model: \xia \cite{xia2019emotion}. We also adapted two models that achieve state of the art performance on the ECPE task: \dingone \cite{ding2020ecpe} and \dingtwoisml \cite{ding2020end} and compared them against our model: \archi. The model is trained for 15 epochs using Adam optimizer \cite{kingma2014adam}. The learning rate and batch size were set to 0.005 and 32 respectively. Model weights and biases were initialized by sampling from a uniform distribution ~$\text{U}(-0.10, 0.10)$. GloVe word embeddings \cite{pennington2014glove} of 200 dimension are used. For regularization, we set the dropout rate to 0.8 for word embeddings and L2 weight decay of 1e-5 over softmax parameters. We set $\lambda_{c}:\lambda_{e}:\lambda_{p} =  1:1:2.5$. The values — chosen through grid search on the hyperparameter space — reflect the higher importance of the primary pair detection task compared to the auxiliary tasks. To obtain better positional embeddings which encode the relative positioning between clauses, we trained randomly initialized embeddings after setting the clipping distance \cite{shaw2018self} to 10 — all clauses that have a distance of 10 or more between them have the same positional embedding. 

We use the same evaluation metrics (precision, recall and F1-score), as used in the past work on ECE and ECPE tasks. 
Following \newcite{xia2019emotion}, the metric definitions are defined as:

\begin{align*}
P  & = \frac{\#correct\_pairs}{\#proposed\_pairs} \hspace{1cm}\\
R  & = \frac{\#correct\_pairs}{\#annotated\_pairs} \hspace{1cm}\\
F1  & = \frac{2*P*R}{P+R}
\end{align*}
where,
\begin{align*}
\#proposed\_pairs =& \text{ no. of emotion-cause} \\ &\text{pairs } \text{predicted by model}\\
\#correct\_pairs =& \text{ number of emotion-cause}\\
& \text{pairs predicted \textit{correctly}}\\ &\text{by the model}\\
\#annotated\_pairs =& \text{ total number of actual}\\
& \text{emotion-cause pairs in} \\ &\text{the data}
\end{align*}

$P,R,F1$ for the two auxiliary classification tasks (emotion-detection and cause-detection) have the usual definition (see \citet{gui-etal-2016-event}).

The results are summarized in Table \ref{tab:results}. Our model \archi outperforms \xia on the task of emotion-cause pair extraction by a significant margin of 6.5\%. This explicitly demonstrates that an end-to-end model works much better since it leverages the mutual dependence between the emotion and cause clauses. As shown in Table \ref{tab:results}, our model achieves comparable performance to the highly parameterized and complex models \dingone and \dingtwoisml (which either leverage a pre-trained BERT \cite{devlin-etal-2019-bert} and 2D Transformer \cite{vaswani2017attention} or iterative BiLSTM encoder). 
To further demonstrate this point, we compare the number of trainable parameters across models in Table \ref{trainable_params}.

\begin{table}[t]
\centering
\begin{tabular}{l*{2}c}
\toprule
Method & Trainable parameters  \\
\hline
\archi & 790,257\\
\dingone & 1,064,886\\
\dingtwoisml & 6,370,452\\
\bottomrule
\end{tabular}
\caption{Comparison of trainable parameters of our model (\archi) with the state-of-the-art models (\dingone and \dingtwoisml). We achieve comparable performance with these models with fewer  parameters and simpler architecture.}
\label{trainable_params}
\end{table}

We also evaluate our model on the ECE task. For this, we use a variant of \archi i.e. \bound, so that it utilizes the emotion annotations. Specifically, we incorporate the knowledge of emotion annotations by incorporating them into the \texttt{Cause-Encoder} as well as the \texttt{Pair-Prediction-Module} and show that this improves performance in both the primary pair prediction task as well as the auxiliary cause detection task (appendix section \ref{sup:variants}). 
\bound outperforms the state of the art model \rhnn: \cite{fan-etal-2019-knowledge} on the ECE task. This is indicative of the generalization capability of our model and further demonstrates that performance on ECPE can be enhanced with improvements in the quality of emotion predictions.

\subsection{Ablation Experiments}
We analyzed the effect of different components of the model on performance via ablation experiments. We present some notable results below. More results are presented in the Appendix.

\noindent\textbf{Positional Embeddings:} 
For finding the extent to which positional embeddings affect the performance, we train \archi without positional embeddings. This resulted in a slight drop in validation F1 score from 51.34 to 50.74, which suggests that the network is robust to withstanding the loss of distance information at the clause-level.

\begin{figure}[h]
\centering
\includegraphics[width=1\linewidth]{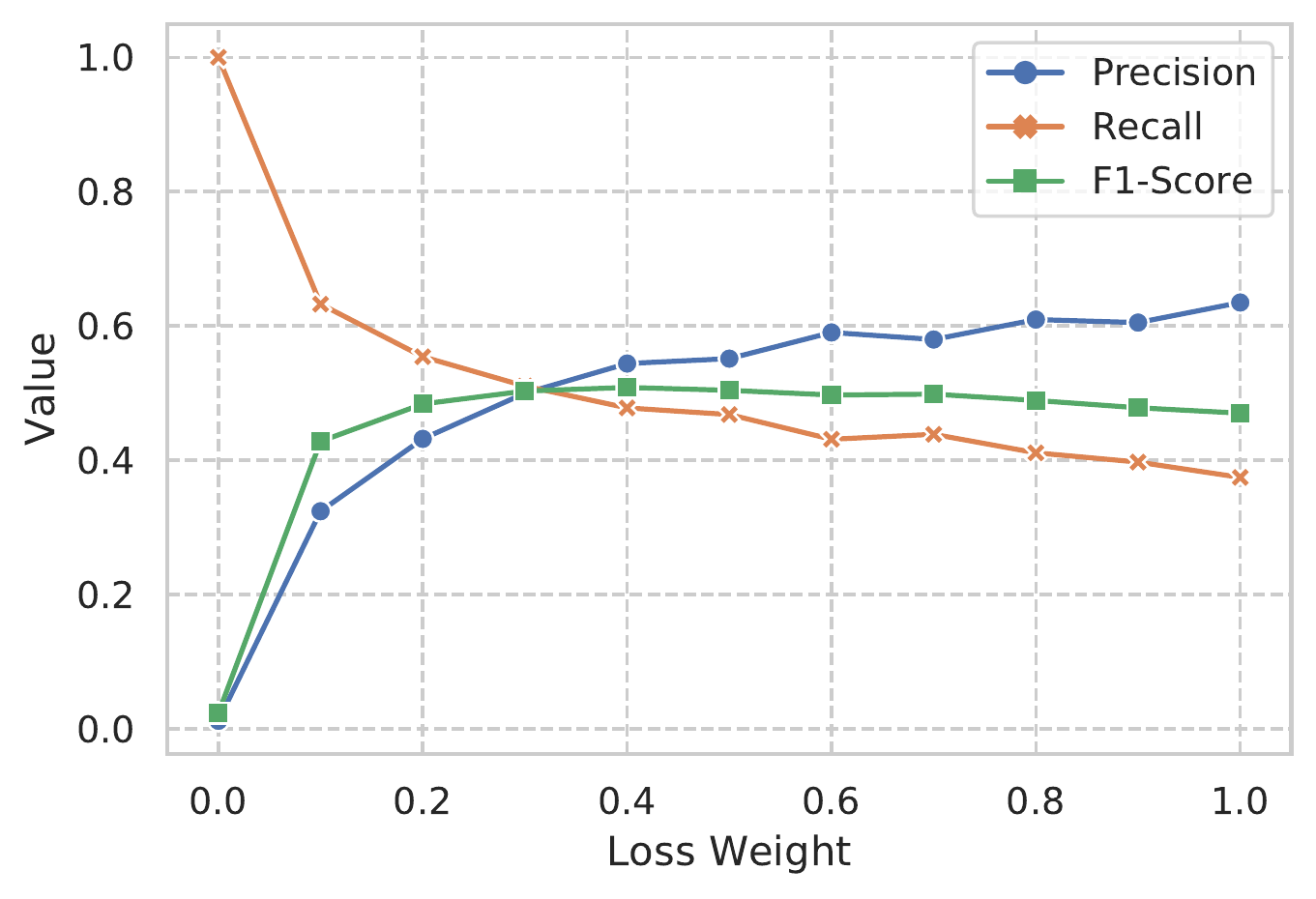}
\caption{Precision, Recall and F1 Score as a function of weight assigned to negative examples.}
\label{fig:losswgt}
\end{figure}

\noindent\textbf{Loss Weighting:} To handle the problem of data imbalance, we varied loss weight. We observed (Figure \ref{fig:losswgt}) that assigning less weight to the negative examples leads to more predicted positives, and hence a better recall but worse precision.

\section{Conclusion}
In this paper, we demonstrated that a simple end-to-end model can achieve competitive performance on the ECPE task by leveraging the inherent correlation between emotions and their causes and optimizing directly on the overall objective. We also showed that a variant of our model which further uses emotion annotations, outperforms the previously best performing model on the ECE task, thereby showing its applicability to variety of related tasks involving emotion analysis. In future, we plan on developing a larger benchmark English-language dataset for the ECPE task. We plan to explore other model-architectures which we expect will help us learn richer representations for causality detection in clause-pairs.

\bibliography{eacl2021}
\bibliographystyle{acl_natbib}


\appendix
\begin{figure*}[t]
\centering
\includegraphics[scale=0.75]{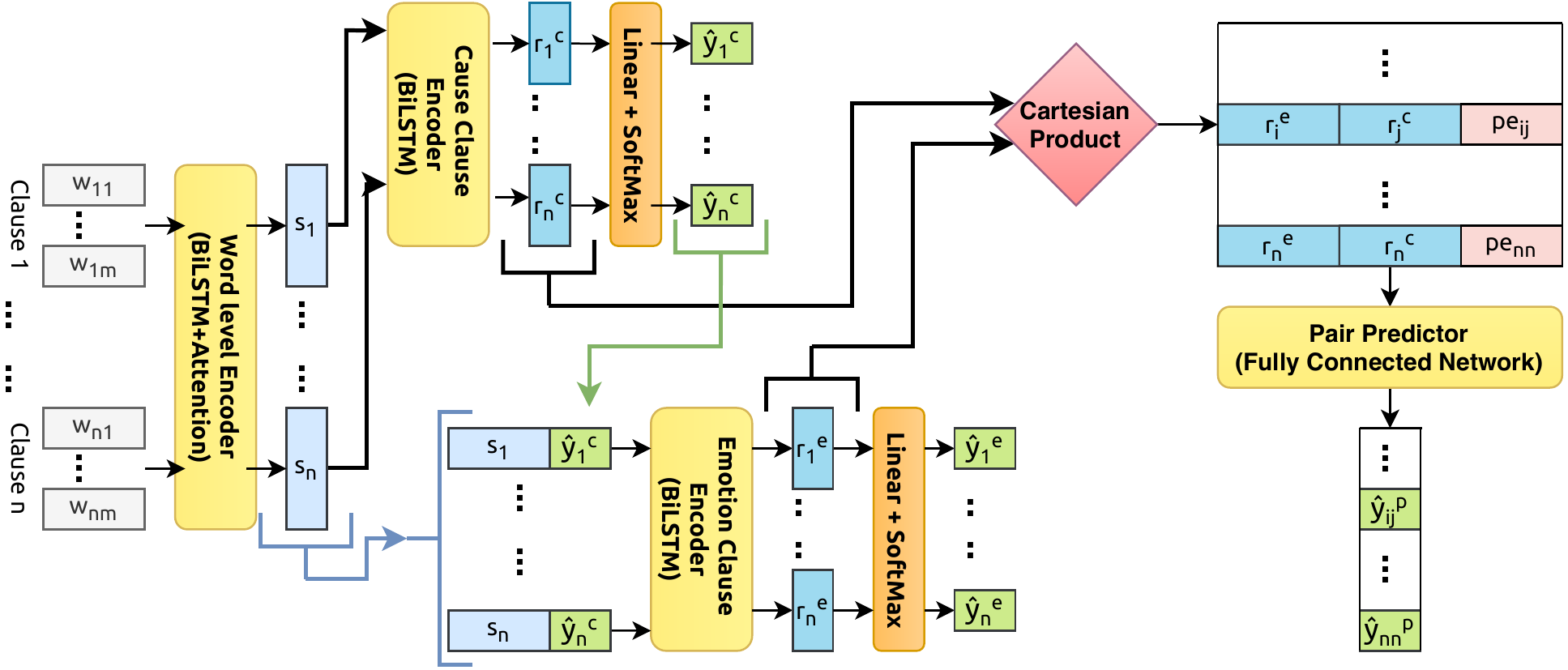}
\caption{End-to-End network variant (\ce)}
\label{fig:E2E2}
\end{figure*}

\section*{{\Large\selectfont{Appendix}}}
\label{supp}
\section{Model Variants} \label{sup:variants}
We present a variant of \archi called \ce (Figure \ref{fig:E2E2}). Here, we feed the clause embeddings $\mathbf{s}_{i}$ into a clause-level BiLSTM to obtain context-aware clause encodings $\mathbf{r}_{i}^{c}$ which is fed into a softmax layer to obtain cause predictions $\mathbf{y}_{i}^{c}$. We concatenate the cause predictions with the clause embeddings, $\mathbf{s}_{i} \oplus \mathbf{y}_{i}^{c}$, and feed them into another clause-level BiLSTM to obtain emotion representations $\mathbf{r}_{i}^{e}$, which are fed into another softmax layer to obtain the emotion predictions $\mathbf{y}_{i}^{e}$. The pair prediction network remains the same as described for \archi.  For finding the extent to which emotion labels can help in improving pair predictions, we present a variant of \archi called \bound. We use the true emotion labels instead of emotion predictions $\mathbf{y}_{i}^{e}$ to obtain the context-aware clause encodings $\mathbf{r}_{i}^{c}$. We also concatenate them with the input of pair-prediction network $\mathbf{r}_{i}^{e} \oplus \mathbf{r}_{j}^{c} \oplus \mathbf{pe}_{ij}$ to make full use of the additional knowledge of emotion labels. Similarly, the corresponding variant of \ce which utilizes true cause labels is called \cebound. The results of model variation are shown in Table \ref{tab:results2}. Here, the pair prediction network consists of a single fully connected layer. After comparing the performance of \ce and \cebound we conclude that huge improvements in performance can be achieved if the quality of cause predictions is improved.

\begin{table*}[h]
\small
\centering
\label{results table}
\resizebox{\linewidth}{!}{
\begin{tabular}{@{} l rrr rrr rrr @{} }
\toprule
 & \multicolumn{3}{c}{Emotion Extraction} & \multicolumn{3}{c}{Cause Extraction} & \multicolumn{3}{c}{Pair Extraction}\\ \cmidrule(lr){2-4} \cmidrule(lr){5-7} \cmidrule(l{0mm}r){8-10} 
 & Precision & Recall& F1 Score & Precision & Recall& F1 Score & Precision & Recall& F1 Score \\
\midrule
 \ce & 71.70 & 66.77 & 69.10 & 63.75 & 42.50 & 50.42 &  48.88 & 48.22 & 48.37\\
 \cebound & 81.74 & 71.95 & 76.47 & 100.00 & 100.00 & 100.00 & 74.17 & 78.63 & 76.30\\
\bottomrule
\end{tabular}
}
\caption{Results of Model variants using precision, recall, and F1-score on the ECPE task and the two sub-tasks.}
\label{tab:results2}
\end{table*}


\end{document}